\documentclass[10pt,preprint]{sigplanconf}

\usepackage{amsmath, amssymb}

\usepackage{xspace}
\usepackage{balance}
\usepackage{url}
\usepackage{comment}
\usepackage{amsmath, amssymb}
\usepackage{latexsym}      
\usepackage{graphics}
\usepackage{subfigure}
\usepackage{epsfig}
\usepackage{epstopdf}
\usepackage{microtype}
\usepackage{flushend}
\usepackage{placeins}
\usepackage{datetime}
\usepackage[usenames,dvipsnames,svgnames,table]{xcolor}
\usepackage{paralist}
\usepackage{url}
\usepackage{hyperref}
\usepackage{multirow}

\usepackage{listings}
\definecolor{darkblue}{rgb}{0,0,.6}
\definecolor{darkred}{rgb}{.6,0,0}
\definecolor{darkgreen}{rgb}{0,.6,0}
\definecolor{red}{rgb}{.98,0,0}
\usepackage{algorithm}
\usepackage{algpseudocode}

\newcommand{\tabincell}[2]{\begin{tabular}{@{}#1@{}}#2\end{tabular}}
\newcommand{\system}{\mbox{MLtuner}\xspace}
\newcommand{\System}{\mbox{MLtuner}\xspace}
\newcommand{\nfx}{\tilde{x}^{(nf)}}

\newif\ifcommenton
\ifcommenton
\newcommand{\greg}[1]{{\bf\color{red} [Greg: #1]}}
\newcommand{\cui}[1]{{\bf\color{blue} [Cui: #1]}}
\newcommand{\phil}[1]{{\bf\color{green} [Phil: #1]}}
\else
\newcommand{\greg}[1]{}
\newcommand{\cui}[1]{}
\newcommand{\phil}[1]{}
\fi

\date{}

\begin{document}

\title{\Large \bf \System: System Support for Automatic Machine Learning Tuning}

\authorinfo{Henggang Cui, Gregory R. Ganger, Phillip B. Gibbons}
           {Carnegie Mellon University}

\maketitle

\begin{abstract}
\system automatically tunes settings for training tunables---such as
the learning rate, the momentum, the mini-batch size,
and the data staleness bound---that
have a significant impact on large-scale machine learning (ML)
performance.
Traditionally, these tunables are set manually,
which is unsurprisingly error prone and difficult to do
without extensive domain knowledge.
\system uses efficient snapshotting, branching, and optimization-guided online trial-and-error
to find good initial settings as well as to re-tune settings during
execution.
Experiments show that
\System can robustly find and re-tune tunable settings
for a variety of ML applications,
including image classification (for 3~models and 2~datasets),
video classification, and matrix factorization.
Compared to state-of-the-art ML auto-tuning approaches,
\system is more robust for large problems and over an order of magnitude
faster.

\end{abstract}

%
%

\section{Introduction}
\label{sec:intro}

Large-scale machine learning (ML) is quickly becoming a common
activity for many organizations.
ML training computations generally use iterative algorithms to converge
on thousands to millions of parameter values that make a pre-chosen
model (e.g., a neural network or pair of factor matrices)
statistically approximate the reality corresponding to the input
training data over which they iterate.
Trained models can be used to predict, cluster, or otherwise help explain
subsequent data.

For training of large, complex models, parallel execution over multiple cluster
nodes is warranted.
The algorithms and frameworks used generally have multiple tunables
that have significant impact on the execution and convergence rates.
For example, the learning rate is a key tunable when using stochastic
gradient descent (SGD) for training.
As another example, the data staleness bound is a key tunable when using
frameworks that explicitly balance asynchrony benefits with inconsistency
costs~\cite{lazytable,muli-ps}.

These tunables are usually manually configured or left to broad defaults.
Unfortunately, knowing the right settings is often quite challenging.
The best tunable settings can depend on the chosen model,
model hyperparameters (e.g., number and types of layers in a neural network),
the algorithm, the framework, and the resources on which the ML application
executes.
As a result, manual approaches not surprisingly involve considerable
effort by domain experts
or yield (often highly) suboptimal training times and solution quality.
Our interactions with both experienced and novice ML users comport with
this characterization.

\system is a tool for automatically tuning ML application training tunables.
It hooks into and guides a training system in trying different settings.
\system determines initial tunable settings based on rapid trial-and-error
search, wherein each option tested runs for a small (automatically determined)
amount of time, to
find good settings based on the convergence rate.
It repeats this process when convergence slows, to see if different
settings provide faster convergence and/or better solution.
This paper describes \system{}'s design and how it addresses
challenges in auto-tuning ML applications, such as
large search spaces, noisy convergence progress, 
variations in effective trial times,
best tunable settings changing over time, when to re-tune, etc.
\cui{When to re-tune is not very interesting any more.}

We have integrated \system with two different state-of-the-art
training systems and experimented with several real ML applications,
including a recommendation application on a CPU-based
parameter server system and both image classification and video
classification on a
GPU-based parameter server system.
For increased breadth, we also experimented with three different popular
models and two datasets for image classification.
The results show \system{}'s effectiveness:
\system consistently zeroes in on good tunables, in each case,
guiding training to match and exceed the best settings we have found
for the given application/model/dataset.
Comparing to state-of-the-art hyperparameter tuning approaches,
such as Spearmint~\cite{spearmint} and Hyperband~\cite{hyperband},
\system completes over an order of magnitude faster and does not
exhibit the same robustness issues for large models/datasets.

This paper makes the following primary contributions.  First, it
introduces the first approach for automatically tuning the multiple
tunables associated with an ML application within the context
of a single execution of that application.  Second, it describes a
tool (\system) that implements the approach, overcoming various
challenges, and how \system was integrated with two different ML
training systems.  Third, it presents results from experiments with
real ML applications, including several models and datasets,
demonstrating the efficacy of this new approach
in removing the ``black art'' of tuning from ML application training
without the orders of magnitude runtime increases of existing auto-tuning
approaches.

\section{Background and related work}
\label{sec:background}

\subsection{Distributed machine learning}

The goal of an ML task is to
train the \emph{model parameters} of an ML model on a set of training data,
so that the trained model can be used
to make predictions on unseen data.
The fitness error of the model parameters to the training data is
defined as the \emph{training loss},
computed from an \emph{objective function}.
The ML task often minimizes the objective function (thus the training loss)
with an iterative convergent algorithm,
such as stochastic gradient descent (SGD).
The model parameters are first initialized randomly,
and in every step, the SGD algorithm samples one \emph{mini-batch}
of the training data
and computes the gradients of the objective function,
w.r.t.\ the model parameters.
The parameter updates will be the opposite direction of the gradients,
multiplied by a \emph{learning rate}.

To speed up ML tasks,
users often distribute the ML computations
with a \emph{data parallel} approach,
where the training data is partitioned across multiple ML workers.
Each ML worker keeps a local copy of the model parameters
and computes parameter updates based on
its training data and local parameter copy.
The ML workers propagate their parameter updates
and refresh their local parameter copies with the updates
every \emph{clock},
which is often logically defined as some quantity of work
(e.g., every training data batch).
Data parallel training is often achieved with
a \emph{parameter server} system~\cite{muli-ps,geeps,bosen,iterstore,
projectadam,dean2012large,yahoolda,piccolo},
which manages a global version of the parameter data
and aggregates the parameter updates from the workers.


\subsection{Machine learning tunables}

ML training often requires the selection and tuning of
many \emph{training hyperparameters}.
For example, the SGD algorithm has a \emph{learning rate} (a.k.a.\ step size)
hyperparameter that controls the magnitude of the model parameter updates.
The \emph{training batch size} hyperparameter controls
the size of the training data mini-batch that each worker processes each clock.
Many deep learning applications use the momentum
technique~\cite{sutskever2013importance} with SGD,
which exhibit a \emph{momentum} hyperparameter,
to smooth updates across different training batches.
In data-parallel training,
ML workers can have temporarily inconsistent parameter copies,
and in order to guarantee model convergence,
consistency models (such as SSP~\cite{ssp,lazytable}
or bounded staleness~\cite{muli-ps})
are often used,
which provide tunable \emph{data staleness} bounds.

Many practitioners (as well as our own experiments)
have found that the settings of the training hyperparameters
have a big impact on
the completion time of an ML task
(e.g., orders of magnitude slower with bad settings)
~\cite{senior2013empirical,ngiam2011optimization,
adagrad,adarevision,adadelta,adam,ssp,lazytable}
and even the quality of the converged model
(e.g., lower classification accuracy with bad settings)
~\cite{senior2013empirical,ngiam2011optimization}.
To emphasize that training hyperparameters need to be tuned,
we call them \emph{training tunables} in this paper.

The training tunables should be distinguished
from another class of ML hyperparameters,
called \emph{model hyperparameters}.
The training tunables control the training procedure
but do not change the model (i.e., they are not in the objective function),
whereas the model hyperparameters define the model
and appear in the objective function.
Example model hyperparameters include
model type (e.g., logistic regression or SVM),
neural network depth and width,
and regularization method and magnitude.
\System focuses on improving the efficiency of training tunable tuning,
and could potentially be used to select training tunables
in an inner loop of existing approaches that tune
model hyperparameters.


\subsection{Related work on machine learning tuning}


\subsubsection{Manual tuning by domain experts}

The most common tuning approach is to do it manually
(e.g.,~\cite{inception-v3, inception-bn, googlenet, alexnet, resnet, poseidon}).
Practitioners often either use some
uilt-in defaults or
pick training tunable settings via trial-and-error.
Manual tuning is inefficient,
and the tunable settings chosen are often suboptimal.

For some tasks, such as training a deep neural network for image classification,
practitioners find it is important to decrease the learning rate during training
in order to get a model with good classification accuracy~\cite{
inception-v3, inception-bn, googlenet, alexnet, resnet, poseidon},
and there are typically two approaches of doing that.
The first approach
(taken by~\cite{alexnet, poseidon})
is to manually change the learning rate
when the classification accuracy plateaus (i.e., stops increasing),
which requires considerable user efforts for monitoring the training.
The second approach
(taken by~\cite{inception-v3, inception-bn, googlenet})
is to decay the learning rate $\eta$ over time $t$,
with a function of $\eta = \eta_0 \times \gamma^t$.
The learning rate decaying factor $\gamma$,
as a training tunable, is even harder to tune than the learning rate,
because it affects the future learning rate.
To decide the best $\gamma$ setting for a training task,
practitioners often need to train the model to completion several times,
with different $\gamma$ settings.

\subsubsection{Traditional hyperparameter tuning approaches}

There is prior work on automatic ML hyperparameter tuning
(sometimes also called model search),
including~\cite{spearmint, hyperband, tupaq,
vartaksupporting, bergstra2013making, komer2014hyperopt, bergstra2011algorithms,
maclaurin2015gradient, pedregosa2016hyperparameter,
feurer2015efficient, thornton2013auto}.
However, none of the prior work
distinguishes training tunables from model hyperparameters;
instead, they tune both of them together in a combined search space.
Because many of their design choices are made for
model hyperparameter tuning,
we find them inefficient and insufficient for training tunable tuning.

\cui{``traditional'' or ``existing''?}
To find good model hyperparameters,
many traditional tuning approaches
train models \emph{from initialization to completion}
with different hyperparameter settings
and pick the model with the best quality
(e.g., in terms of its cross-validation accuracy for a classification task).
The hyperparameter settings to be evaluated are often decided with
\emph{bandit optimization algorithms},
such as Bayesian optimization~\cite{movckus1975bayesian}
or \mbox{HyperOpt}~\cite{hyperopt}.
Some tuning approaches,
such as Hyperband~\cite{hyperband} and TuPAQ~\cite{tupaq},
reduce the tuning cost
by stopping the lower-performing settings early,
based on the model qualities achieved in the early stage of training.

\System differs from existing approaches in several ways.
First, \system trains the model to completion only once,
with the automatically decided best tunable settings,
because training tunables do not change the model,
whereas existing approaches
train multiple models to completion multiple times,
incurring considerable tuning cost.
Second, \system uses training loss,
rather than cross-validation model qualities,
as the feedback to evaluate tunable settings.
Training loss can be obtained for every training batch at no extra cost,
because SGD-based training algorithms use training loss
to compute parameter updates,
whereas the model quality is evaluated
by testing the model on validation data,
and the associated cost does not allow it to be frequently evaluated
(often every thousands of training batches).
Hence, \system can use more frequent feedback to find good tunable
settings in less time
than traditional approaches.
This option is enabled by the fact that, unlike model hyperparameters,
training tunables
do not change the mathematical formula of the objective function,
so just comparing the training loss is sufficient.
Third, \system automatically decides
the amount of resource (i.e., training time) to use
for evaluating each tunable setting,
based on the noisiness of the training loss,
while existing approaches either hard-code the trial effort
(e.g., TuPAQ always uses 10~iterations)
or decide it via a grid search
(e.g., Hyperband iterates over each of the possible resource allocation plans).
Fourth, \system is able to re-tune tunables during training,
while existing approaches
use the same hyperparameter setting for the whole training.
Unlike model hyperparameters, training tunables can (and often should)
be dynamically changed during training, as discussed above.


\subsubsection{Adaptive SGD learning rate tuning algorithms}
\label{sec:background-sgd-lr-tuning}

Because the SGD algorithm is well-known for
being sensitive to the learning rate (LR) setting,
experts have designed many adaptive SGD learning rate tuning algorithms,
including \mbox{AdaRevision}~\cite{adarevision}, \mbox{RMSProp}~\cite{rmsprop},
\mbox{Nesterov}~\cite{nesterov}, \mbox{Adam}~\cite{adam},
\mbox{AdaDelta}~\cite{adadelta}, and \mbox{AdaGrad}~\cite{adagrad}.
These algorithms adaptively adjust the LR
for individual model parameters based on the magnitude of their gradients.
For example, they often use relatively large LRs for
parameters with small gradients
and relatively small LRs
for parameters with large gradients.
However, these algorithms still require users
to set the initial LR.
Even though they are less sensitive
to the initial LR settings
than the original SGD algorithm,
our experiment results in Section~\ref{sec:eval-sgd-lr-tuning}
show that bad initial LR settings can
cause the training time to be orders of magnitude longer
(e.g., Figure~\ref{fig:all-lrs-mf})
and/or cause the model to converge to suboptimal solutions
(e.g., Figure~\ref{fig:all-lrs-cifar}).
Hence, \system complements
these adaptive LR algorithms,
in that users can use \system to pick the initial LR
for more robust performance.
Moreover, practitioners also find that
sometimes using these adaptive LR tuning algorithms alone
is not enough to achieve the optimal model solution,
especially for complex models such as deep neural networks.
For example, Szegedy et al.~\cite{inception-v3}
reported that they used LR decaying together with \mbox{RMSProp}
to train their \mbox{Inception}-v3 model.

\nocite{tensorflow}

\section{\System: more efficient automatic tuning}
\label{sec:design}


This section describes
the high level design of our \emph{\system} approach.

\subsection{\System overview}

\System automatically tunes training tunables with low overhead,
and will dynamically re-tune tunables during the training.
\System is a light-weight system that
can be connected to existing training systems,
such as a parameter server.
\System sends the tunable setting trial instructions to the training system
and receives training feedback (e.g., per-clock training losses)
from the training system.
The detailed training system interfaces
will be described in Section~\ref{sec:impl-interface}.
Similar to the other hyperparameter tuning approaches,
such as Spearmint~\cite{spearmint}, Hyperband~\cite{hyperband},
and TuPAQ~\cite{tupaq},
\system requires users to specify the tunables to be tuned,
with the type---either discrete,
continuous in linear scale, or continuous in log scale---and range of valid values.

\subsection{Trying and evaluating tunable settings}

\begin{figure}[ht]
\centering
\includegraphics[keepaspectratio=1,width=1\columnwidth]{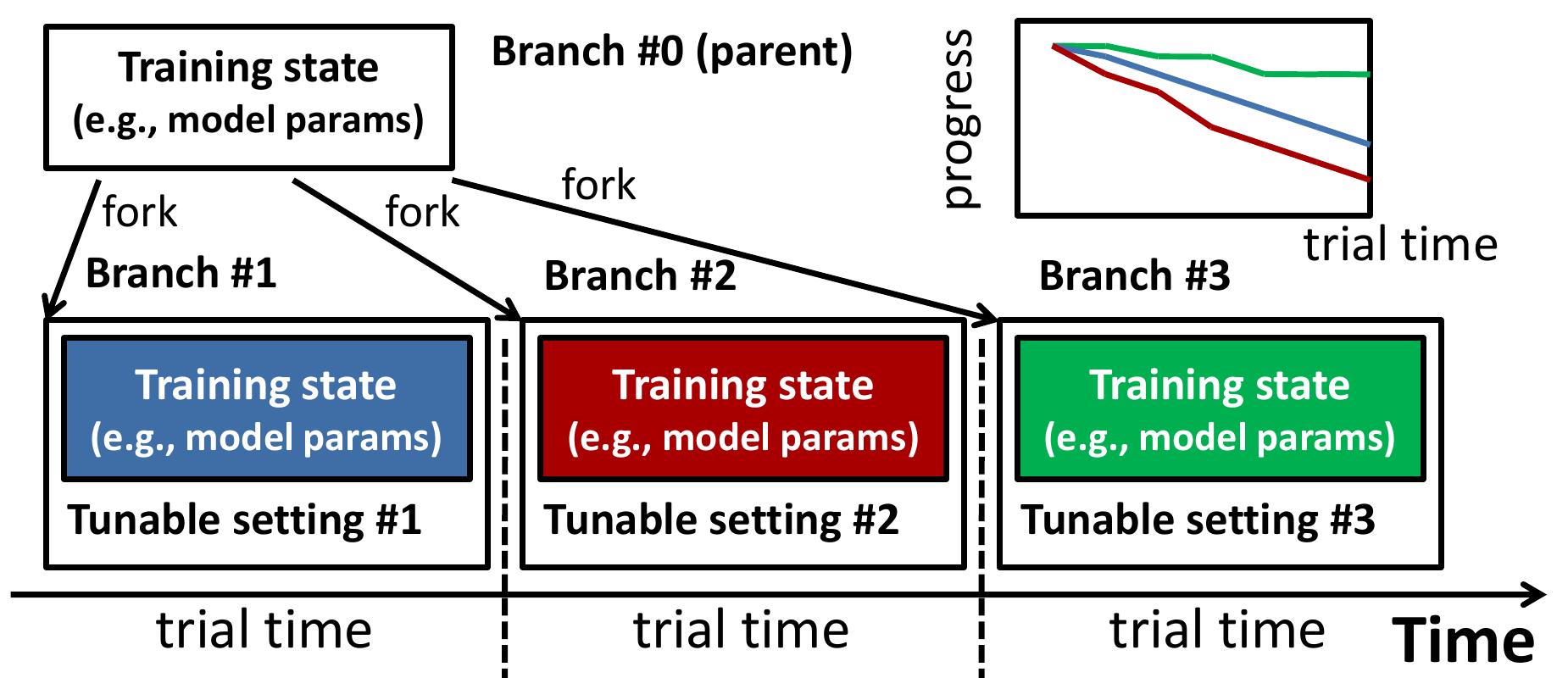}
\vspace{-.1in}
\caption{Trying tunable settings in training branches.
\small{The red branch with tunable setting \#2 has the fastest convergence speed.}}
\label{fig:training-branches}
\end{figure}

\System evaluates tunable settings
by trying them in forked \emph{training branches}.
The training branches are forked from the same consistent snapshot
of some initial training state
(e.g., model parameters, worker-local state, and training data),
but are assigned with different tunable settings to train the model.
As is illustrated in Figure~\ref{fig:training-branches},
\system schedules each training branch to run
for some automatically decided amount of \emph{trial time},
and collects their \emph{training progress}
to measure their \emph{convergence speed}.
\system will fork multiple branches to try different settings,
and then pick only the branch with the fastest convergence speed
to keep training, and kill the other branches.
In our example applications,
such as deep learning and matrix factorization,
the training system reports the per-clock training losses to \system
as the training progress.

The training branches are scheduled by \system in a \emph{time-sharing} manner,
running in the same training system instance on the same set of machines.
We made this design choice,
rather than running multiple training branches in parallel
on different sets of machines,
for three reasons.
First, this design avoids the use of extra machines
that are just for the trials;
otherwise, the extra machines will be wasted
when the trials are not running (which is most of the time).
Second, this design allows us to use the same hardware setup
(e.g., number of machines)
for both the tuning and the actual training;
otherwise, the setting found on a different hardware setup
would be suboptimal for the actual training.
Third, running all branches in the same training system instance
helps us achieve low overhead for
forking and switching between branches,
which are now simply memory copying of training state within the same process
and choosing the right copy to use.
Also, some resources, such as cache memory and immutable training data,
can be shared among branches without duplication.


\subsection{Tunable tuning procedure}
\label{sec:design-tunable-searching}

\begin{figure}[ht]
\centering
\includegraphics[keepaspectratio=1,width=1\columnwidth]{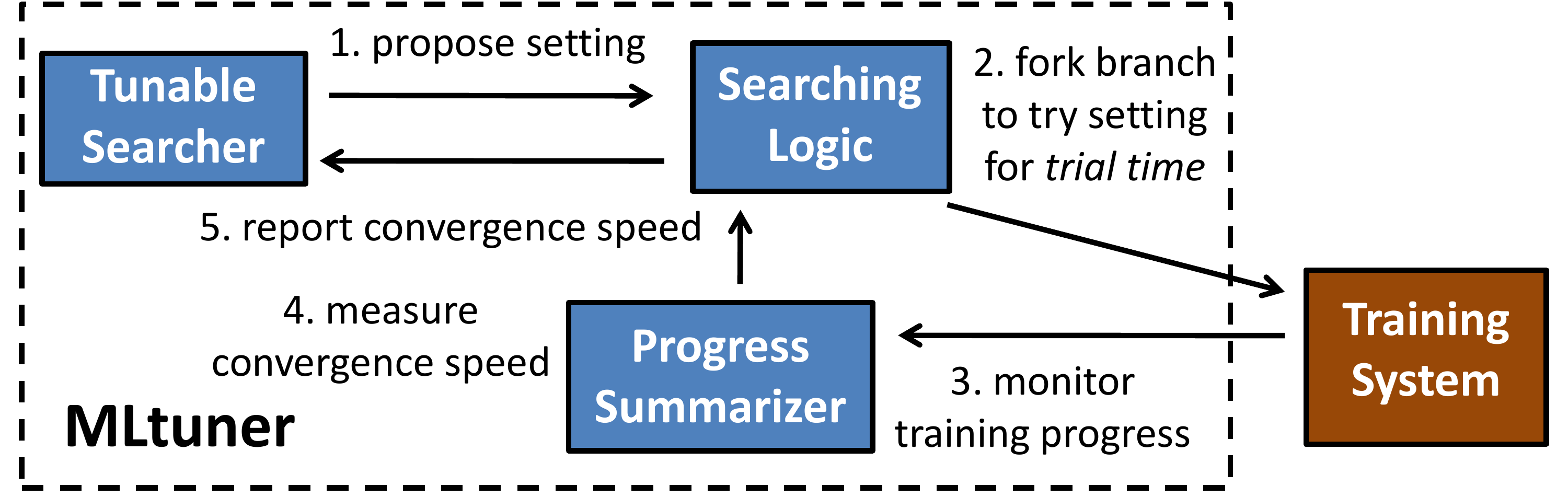}
\vspace{-.1in}
\caption{\System tuning procedure.}
\label{fig:tunable-search-procedure}
\end{figure}

Figure~\ref{fig:tunable-search-procedure} illustrates
the tuning procedure of \system.
\System first tags the current training state
as the parent branch.
Then, inside the tuning loop,
\system uses a \emph{tunable searcher} module
(described in Section~\ref{sec:impl:tunable-searcher})
to decide the next tunable setting to evaluate.
For each proposed trial setting,
\system will instruct the training system to
fork a trial branch from the parent branch
to train the model for some amount of \emph{trial time}
with the trial setting.
Section~\ref{sec:impl:decide-trial-time}
will describe how \system automatically decides the trial time.
Then, \system will collect the training progress of the trial branch
from the training system,
and use the \emph{progress summarizer} module
(described in Section~\ref{sec:impl:summarize-progress})
to summarize its \emph{convergence speed}.
The convergence speed will be reported back to the tunable searcher
to guide its future tunable setting proposals.
\System uses this tuning procedure to
tune tunables at the beginning of the training,
as well as to re-tune tunables during training.
Re-tuning will be described in Section~\ref{sec:impl:retune}.

\subsection{Assumptions and limitations}
\label{sec:design-limitations}

The design of \system relies on the assumption that the good tunable settings
(in terms of completion time and converged model quality)
can be decided based on their convergence speeds
measured with a relatively short period of trial time.
The same assumption has also been made by
many of the state-of-the-art hyperparameter tuning approaches.
For example, both Hyperband and TuPAQ stop some of the
trial hyperparameter settings early,
based on the model qualities achieved in the early stage of the training.
This assumption does not always hold for model hyperparameter tuning.
For example, a more complex model often takes more time to converge
but will eventually converge to a better solution.
For most of the training tunables,
we find this assumption holds for all the applications
that we have experimented with so far,
including image classification on two different datasets
with three different deep neural networks,
video classification, and matrix factorization.
That is because the training tunables only control the training procedure
but do not change the model.
That is also the reason why we do not suggest using
\system to tune the model hyperparameters.
For use cases where
both training tunables and model hyperparameters need to be tuned,
users can use \system in the inner loop to tune the training tunables,
and use the existing model hyperparameter tuning approaches in the outer loop
to tune the model hyperparameters.

\section{\system implementation details}
\label{sec:impl}

This section describes the design and implementation details of \system.

\subsection{Measuring convergence speed}
\label{sec:impl:summarize-progress}

The progress summarizer module takes in the training progress trace
(e.g., a series of training loss) of each trial branch,
and outputs the convergence speed.
The progress trace has the form of $\{(t_i, x_i)\}_{i=1}^N$,
where $t_i$ is the timestamp, and $x_i$ is the progress.
In this section, we assume $x$ is the training loss,
and a smaller $x$ value means better convergence.


\textbf{Downsampling the progress trace.}
The most straightforward way of measuring the convergence speed
is to use the slope of the progress trace: $s=\frac{|x_N-x_1|}{t_N-t_1}$.
However, in many ML applications,
such as deep neural network training with SGD,
the progress trace is often quite noisy,
because the training loss points are computed on
different batches of the training data.
We find the convergence speed measured with just the first and last point
of the progress trace is often inaccurate.
To deal with the noisiness, the progress summarizer will
\emph{downsample} the progress trace of each branch,
by uniformly dividing the progress trace
into $K$ non-overlapping windows.
The value of each window will be calculated as
the average of all data points in it.
For a downsampled progress trace
of $\{(\tilde{t}_i, \tilde{x}_i)\}_{i=1}^K$,
its slope will be $\tilde{s}=\frac{-range(\tilde{x})}{range(\tilde{t})}$,
where $range(\tilde{x}) = \tilde{x}_K - \tilde{x}_1$
and $range(\tilde{t}) = \tilde{t}_K - \tilde{t}_1$.
We will describe how we decide $K$ later in this section.




\textbf{Penalizing unstable branches.}
Even with downsampling,
calculating the slope by simply looking at the first and
last downsampled points might still treat branches
with unstable jumpy loss as good converging branches.
To deal with this problem, the summarizer module
will adjust the convergence speed estimation of each branch
according to its noisiness.
Ideally, the loss of a noise-free trace $\nfx$
should be monotonically decreasing,
so we estimate the noisiness of a trace $\tilde{x}$ as
$noise(\tilde{x})=\max{(\max\limits_{1 \leq i \leq K-1}{(\tilde{x}_{i+1} - \tilde{x}_i)}, 0)}$,
which is the maximum magnitude that a point goes up from the previous point.
In order to make a conservative estimation of convergence speeds,
our progress summarizer will penalize the convergence speed
of each branch with its noise: \\
$speed=\max{(\frac{-range(\tilde{x})-noise(\tilde{x})}{range(\tilde{t})}, 0)}$.
which is a positive value for a converging branch
and zero for a diverged branch.
We report zero as the convergence speed of a diverged branch,
rather than reporting it as a negative value,
because we find
it is usually wrong to treat a diverged branch with smaller diverged loss
as a better branch than other diverged branches.
We treat diverged branches as of the same quality.

\textbf{Convergence and stability checks.}
The progress summarizer will check the convergence and stability
of each branch,
and assign one of the three labels to them:
\texttt{converging}, \texttt{diverged}, or \texttt{unstable}.
It labels a branch as \texttt{converging},
if $range(\tilde{x}) < 0$ and
$noise(\tilde{x}) < \epsilon \times |range(\tilde{x})|$.
We will describe how we decide $\epsilon$ later in this section.
It labels a branch as \texttt{diverged},
if the training encounters numerically overflowed numbers.
Finally, it labels all the other branches as \texttt{unstable},
meaning that their convergence speeds might need longer trials to evaluate.
With a longer trial time,
an unstable branch might become stable,
because its $|range(\tilde{x})|$
is likely to increase because of the longer training,
and its $noise(\tilde{x})$ is likely to decrease
because of more points being averaged in each downsampling window.


\textbf{Deciding number of samples and stability threshold.}
The previously described progress summarizer module has two knobs,
the number of samples $K$ and the stability threshold $\epsilon$.
Since the goal of \system is to free users from tuning,
$K$ and $\epsilon$ do not need to be tuned by users either.
To decide $K$, we will consider an extreme case,
where the true convergence progress $|range(\nfx)|$ is zero,
and $\tilde{x}$ is all white noise, following a normal distribution.
This trace will be falsely labelled as \texttt{converging},
if $\{\tilde{x}_i\}_{i=1}^K$ is monotonically decreasing,
and this probability is less than ${(\frac{1}{2})}^K$.
Hence, we need a large enough $K$ to bound this false positive probability.
We decide to set $K=10$ to counter the noisiness,
so that the false positive probability is less than 0.1\%.
The $\epsilon$ configuration bounds the magnitude
(relative to $|range(\tilde{x})|$)
that each point in the progress trace is
allowed to go up from the previous point.
On average, if we approximate the noise-free progress trace $\nfx$
as a straight line,
each point is expected to go down from
the previous point by $\approx\frac{|range(\tilde{x})|}{K}$.
Hence, \System sets $\epsilon$ to $\frac{1}{K}$,
so that a \texttt{converging} trace
will have no point going up by more than it is expected to go down.
Our experiments in Section~\ref{sec:evaluation}
show that the same settings of $K$ and $\epsilon$ work robustly
for all of our application benchmarks.

\subsection{Deciding tunable trial time}
\label{sec:impl:decide-trial-time}

Unlike traditional tuning approaches,
\system automatically decides tunable trial time,
based on the noisiness of training progress,
so that the trial time is just long enough
for good tunable settings to have stable converging progress.
Algorithm~\ref{alg:decide-trial-time}
illustrates the trial time decision procedure.
\System first initializes the trial time to a small value,
such as making it as long as the decision time of the tunable searcher,
so that the decision time will not dominate.
While \System tries tunable settings,
if none of the settings tried so far is labelled as \texttt{converging}
with the current trial time,
\system will double the trial time
and use the doubled trial time to try new settings
as well as the previously tried settings for longer.
When \system successfully finds a stable converging setting,
the trial time is decided and will be used to evaluate future settings.

\begin{algorithm}
\caption{\label{alg:decide-trial-time}
\System trial time decision.}
\footnotesize
\begin{algorithmic}
\State $trialTime \gets 0$
\State Parent branch $\gets$ current model state
\While{none of the settings is \texttt{converging}}
  \State Get $tunableSetting$ from tunable searcher
  \State $trialTime \gets \max(trialTime, searcherDecisionTime)$
  \If{$tunableSetting$ is not empty}
    \State Fork a branch from the parent branch with $tunableSetting$
    \State Append the new branch to $trialBranches$
  \EndIf
  \For{each $branch$ in $trialBranches$}
  \State Schedule $branch$ to run for $trialTime - branch.runTime$
  \EndFor
  \State Summarize the progress of all $trialBranches$
  \State Remove \texttt{diverged} branches from $trialBranches$
  \If{any $branch$ in $trialBranches$ is \texttt{converging}}
    \State $bestSetting \gets$ tunable setting that has the best convergence
    \State Free the non-best branches
    \State Trial time decided and break out the loop
  \Else
    \State $trialTime \gets trialTime \times 2$
  \EndIf
\EndWhile
\State Keep searching with $trialTime$
\end{algorithmic}
\end{algorithm}

\subsection{Tunable searcher}
\label{sec:impl:tunable-searcher}

The tunable searcher is a replaceable module that
searches for a best tunable setting that maximizes the convergence speed.
It can be modeled as
a black-box function optimization problem (i.e., bandit optimization),
where the function input is a tunable setting,
and the function output is the achieved convergence speed.
\System allows users to choose from
a variety of optimization algorithms,
with a general tunable searcher interface.
In our current implementation, we have implemented and explored four types of
searchers,
including RandomSearcher, GridSearcher,
BayesianOptSearcher, and HyperOptSearcher.

The simplest RandomSearcher just samples
settings uniformly from the search space,
without considering the convergence speeds of previous trials.
GridSearcher is similar to RandomSearcher,
except that it discretizes the continuous search space into a grid,
and proposes each of the discretized settings in the grid.
Despite its simplicity,
we find GridSearcher works surprisingly well
for low-dimensional cases,
such as when there is only one tunable to be searched.
For high-dimensional cases, where there are many tunables to be searched,
we find it is better to use bandit optimization algorithms,
which spend more searching efforts
on the more promising part of the search space.
Our BayesianOptSearcher uses the Bayesian optimization algorithm,
implemented in the Spearmint~\cite{spearmint} package,
and our HyperOptSearcher uses the HyperOpt~\cite{hyperopt} algorithm.
Through our experiments, we find HyperOptSearcher
works best among all the searcher choices for most use cases,
and \system uses it as its default searcher.

The tunable searcher (except for GridSearcher)
need a stopping condition to decide when to stop searching.
Generally, it can just use the default stopping condition
that comes with the optimization packages.
Unfortunately, neither the \mbox{Spearmint} nor \mbox{HyperOpt} package
provides a stopping condition.
They all rely on users to decide when to stop.
After discussing with many experienced ML practitioners,
we used a rule-of-thumb stopping condition
for hyperparameter optimization,
which is to stop searching
when the top five best (non-zero) convergence speeds
differ by less than 10\%.

\subsection{Re-tuning tunables during training}
\label{sec:impl:retune}


\System re-tunes tunables,
when the training stops making further converging progress
(i.e., considered as converged) with the current tunable setting.
We have also explored designs that re-tune tunables more aggressively,
before the converging progress stops,
but we did not choose those designs for two reasons.
First, we find the cost of re-tuning usually
outweighs the increased convergence rate coming from the re-tuned setting.
Second, we find, for some complex deep neural network models,
re-tuning too aggressively might cause them to converge to
suboptimal local minimas.

To re-tune,
the most straightforward approach
is to use exactly the same tuning procedure as is used for initial tuning,
which was our initial design.
However, some practical issues were found, when we deployed it in practice.
For example,
re-tuning happens when the training stops making converging progress,
but, if the training has indeed converged to the optimal solution
and no further converging progress can be achieved with any tunable setting,
the original tuning procedure will
be stuck in the searching loop for ever.

To address this problem, we find it is necessary
to bound both the per-setting trial time and the number of trials
to be performed for each re-tuning.
For the deep learning applications used in Section~\ref{sec:evaluation},
\system will bound the per-setting trial time to be at most one epoch
(i.e., one whole pass over the training data),
and we find, in practice,
this bound usually will not be reached, unless the model has indeed converged.
\System also bounds the number of tunable trials
of each re-tuning
to be no more than the number of trials of the previous re-tuning.
The intuition is that, as more re-tunings are performed,
the likelihood that a better setting is yet to be found decreases.
These two bounds together will guarantee that the searching procedure
can successfully stop for a converged model.

\subsection{Training system interface}
\label{sec:impl-interface}

\begin{table*}
\centering
{\small
  \begin{tabular}{lll}
    \hline
    {\bf Method name} & {\bf Input} & {\bf Description} \\
    \hline
    \multicolumn{3}{c}{\bf Messages sent from \system} \\
    \texttt{ForkBranch} & (clock, branchId, parentBranchId, tunable[, type]) & fork a branch by taking a consistent snapshot at clock \\
    \texttt{FreeBranch} & (clock, branchId) & free a branch at clock \\
    \texttt{ScheduleBranch} & (clock, branchId) & schedule the branch to run at clock \\
    \hline
    \multicolumn{3}{c}{\bf Messages sent to \system} \\
    \texttt{ReportProgress} & (clock, progress) & report per-clock training progress \\
    \hline
\end{tabular}
}
\caption{\system message signatures.}
\label{tab:interface}
\end{table*}

\System works as a separate process that
communicates with the training system via messages.
Table~\ref{tab:interface} lists the message signatures.
\System identifies each branch with a unique \emph{branch ID},
and uses \emph{clock} to indicate logical time.
The clock is unique and totally ordered across all branches.
When \system forks a branch, it expects the training system to
create a new training branch by
taking a consistent snapshot of all state (e.g., model parameters)
from the parent branch
and use the provided tunable setting for the new branch.
When \system frees a branch, the training system can then
reclaim all the resources (e.g., memory for model parameters)
associated with that branch.
\System sends the branch operations in clock order,
and it sends exactly one \texttt{ScheduleBranch} message for every clock.
The training system is expected to report its training progress
with the \texttt{ReportProgress} message every clock.

Although in our \system design,
the branches are scheduled based on time, rather than clocks,
our \system implementation actually sends
the per-clock branch schedules to the training system.
We made this implementation choice,
in order to ease the modification of the training systems.
To make sure that a trial branch runs for (approximately)
the amount of scheduled trial time,
\system will first schedule that branch to run
for some small number of clocks (e.g., three) to measure its per-clock time,
and then schedule it to run for more clocks,
based on the measured per-clock time.
Also,
because \system consumes very few CPU cycles and little network bandwidth,
users do not need to dedicate a separate machine for it.
Instead, users can just run \system on one of the training machines.
\cui{TODO: maybe can move this to the experimental setup.}

\textbf{Distributed training support.}
Large-scale machine learning tasks are often trained
with distributed training systems
(e.g., with a parameter server architecture).
For a distributed training system with multiple training workers,
\system will broadcast the branch operations to all the training workers,
with the operations in the same order.
\system also allows each of the training workers
to report their training progress separately,
and \system will aggregate the training progress
with a user-defined aggregation function.
For all the SGD-based applications in this paper,
where the training progress is the loss computed as the sum of
the training loss from all the workers,
this aggregation function just does the sum.

\textbf{Evaluating the model on validation set.}
For some applications, such as the classification tasks,
the model quality (e.g., classification accuracy)
is often periodically evaluated
on a set of validation data during the training.
This can be easily achieved with the branching support of \system.
To test the model,
\system will fork a branch with a special \texttt{TESTING} flag
as the \emph{branch type},
telling the training system to use this branch
to test the model on the validation data,
and \system will interpret
the reported progress of the testing branch as the validation accuracy.

\subsection{Training system modifications}

This section describes the possible modifications to be made,
for a training system to work with \system.
The modified training system needs to
keep multiple versions of its training state
(e.g., model parameters and local state)
as multiple training branches,
and switch between them during training.

We have modified two state-of-the-art training systems to work with \system:
\mbox{IterStore}~\cite{iterstore}, a generic parameter server system.
and \mbox{GeePS}~\cite{geeps}, a parameter server with specialized
support for GPU deep learning.\footnote{We used the open-sourced
  \mbox{IterStore} code from
\url{https://github.com/cuihenggang/iterstore}
as of November 16, 2016,
and the open-sourced \mbox{GeePS} code from
\url{https://github.com/cuihenggang/geeps}
as of June 3, 2016.}
Both parameter server implementations
keep the parameter data as key-value pairs in memory,
sharded across all worker machines in the cluster.
To make them work with \system,
we modified their parameter data storage modules
to keep multiple versions of the parameter data,
by adding branch ID as an additional field in the index.
When a new branch is forked,
the modified systems will allocate the corresponding data storage for it
(from a user-level memory pool managed by the parameter server)
and copy the data from its parent branch.
When a branch is freed, all its memory will be reclaimed
to the memory pool for future branches.
The extra memory overhead depends on the maximum number of co-existing
active branches,
and \system is designed to keep as few active branches as possible.
Except when exploring the trial time
(with Algorithm~\ref{alg:decide-trial-time}),
\system needs only three active branches to be kept,
the parent branch, the current best branch, and the current trial branch.
Because the parameter server system shards its parameter data
across all machines,
it is usually not an issue to
keep those extra copies of parameter data in memory.
For example, the Inception-BN~\cite{inception-bn} model,
which is the state-of-art convolutional deep neural network
for image classification,
has less than 100~MB of model parameters.
When we train this model on an 8-machine cluster,
the parameter server shard on each machine only needs to
keep 12.5~MB of the parameter data (in CPU memory rather than GPU memory).
A machine with 50~GB of CPU memory
will be able to keep 4000~copies of the parameter data in memory.

Those parameter server implementations also have multiple levels of caches.
For example, both parameter server implementations
cache parameter data locally at each worker machine.
In addition to the machine-level cache,
\mbox{IterStore} also provides a distinct thread-level cache
for each worker thread,
in order to avoid lock contention.
\mbox{GeePS} has a GPU cache that keeps data in GPU memory for GPU computations.
Since \system runs only one branch at a time,
the caches do not need to be duplicated.
Instead, all branches can share the same cache memory,
and the shared caches will be cleared
each time \system switches to a different branch.
In fact, sharing the cache memory is
critical for \mbox{GeePS} to work with \system,
because there is usually not enough GPU memory for \mbox{GeePS} to allocate
multiple GPU caches for different branches.


\begin{table*}
\centering
{\small
  \begin{tabular}{c|c|c|c|c}
    \hline
    {\bf Application} & {\bf Model} & {\bf Supervised/Unsupervised} & {\bf Clock size} & {\bf Hardware} \\
    \hline
    Image classification & \tabincell{c}{Convolutional neural network \\ (\mbox{Inception}-BN, \mbox{GoogLeNet}, \mbox{AlexNet})} & Supervised learning & One mini-batch & GPU \\
    \hline
    Video classification & Recurrent neural network & Supervised learning & One mini-batch & GPU \\
    \hline
    Movie recommendation & Matrix factorization & Unsupervised learning & Whole data pass & CPU \\
    \hline
\end{tabular}
}
\caption{Applications used in the experiments. They have distinct characteristics.}
\label{tab:apps}
\end{table*}

\section{Evaluation}
\label{sec:evaluation}

This section evaluates \system
on several real machine learning benchmarks,
including image classification
with three different models on two different datasets,
video classification, and matrix factorization.
Table~\ref{tab:apps} summarizes the distinct characteristics
of these applications.
The results confirm
that \system can robustly tune and re-tune the tunables
for ML training,
and is over an order of magnitude faster
than state-of-the-art ML tuning approaches.
\cui{Might need to be updated.}

\subsection{Experimental setup}
\label{sec:evaluation:setups}

\subsubsection{Application setup}
\label{sec:evaluation:app-setup}


\textbf{Image classification using convolutional neural networks.}
Image classification is a supervised learning task
that trains a deep convolutional neural network (CNN)
from many labeled training images.
The first layer of neurons (input of the network)
is the raw pixels of the input image,
and the last layer (output of the network)
is the predicted probabilities that the image should be assigned to each
of the labels.
There is a \emph{weight} associated with each neuron connection,
and those weights are the model parameters
that will be trained from the input (training) data.
Deep neural networks are often trained with
the SGD algorithm, which samples one \emph{mini-batch}
of the training data every clock
and computes gradients and parameter updates
based on that mini-batch~\cite{dean2012large,projectadam,geeps,
alexnet,googlenet,inception-bn,inception-v3,resnet}.
As an optimization, gradients are often smoothed across mini-batches
with the \emph{momentum} method~\cite{sutskever2013importance}.

We used two datasets and three models for the image classification experiments.
Most of our experiments used the Large Scale Visual Recognition Challenge 2012
(ILSVRC12) dataset~\cite{ilsvrc12},
which has 1.3~million training images and 5000~validation images,
labeled to 1000~classes.
For this dataset, we experimented with two popular
convolutional neural network models,
\mbox{Inception}-BN~\cite{inception-bn} and
\mbox{GoogLeNet}~\cite{googlenet}.\footnote{The original papers did not release
some minor details of their models,
so we used the open-sourced versions of those models
from the Caffe~\cite{caffe} and MXNet~\cite{mxnet} repositories.}
Some of our experiments also used a smaller
\mbox{Cifar10} dataset~\cite{cifar},
which has 50,000~training images and 10,000~validation images,
labeled to 10~classes.
We used \mbox{AlexNet}~\cite{alexnet} for the \mbox{Cifar10} experiments.

\textbf{Video classification using recurrent neural networks.}
To capture the sequence information of videos,
a video classification task often uses
a \emph{recurrent neural network} (RNN),
and the RNN network is often implemented with
a special type of recurrent neuron layer called
\emph{Long-Short Term Memory} (LSTM)~\cite{hochreiter1997long}
as the building block~\cite{lrcn, vinyals2014show, yue2015beyond}.
A common approach for using RNNs for video classification
is to first encode each image frame of the videos
with a convolutional neural network (such as \mbox{GoogLeNet}),
and then feed the sequences of the encoded image feature vectors
into the LSTM layers.

Our video classification experiments used the UCF-101 dataset~\cite{ucf101},
with about 8,000~training videos and 4,000~testing videos,
categorized into 101~human action classes.
Similar to the approach described by Donahue et al.~\cite{lrcn}
and Cui et al.~\cite{geeps},
we used the \mbox{GoogLeNet}~\cite{googlenet} model,
trained with the ILSVRC12 image data,
to encode the image frames,
and fed the feature vector sequences into the LSTM layers.
We extracted the video frames at a rate of 30~frames per second
and trained the LSTM layers
with randomly selected video clips of 32~frames each.

\textbf{Movie recommendation using matrix factorization.}
The movie recommendation task tries to predict
unknown user-movie ratings, based on a collection of known ratings.
This task is often modeled as a
\emph{sparse matrix factorization} problem,
where we have a partially filled matrix $X$,
with entry $(i, j)$ being user $i$'s rating of movie $j$,
and we want to factorize $X$ into two low ranked matrices $L$ and $R$,
such that their product approximates $X$ (i.e., $X\approx L \times R$)
~\cite{gemulla2011large}.
The matrix factorization model is often trained with
the SGD algorithm~\cite{gemulla2011large},
and because the model parameter values are updated with uneven frequency,
practitioners often use \mbox{AdaGrad}~\cite{adagrad} or \mbox{AdaRevision}~\cite{adarevision}
to adaptively decide the per-parameter learning rate adjustment
from a specified initial learning rate~\cite{bosen}.
Our matrix factorization (MF) experiments used the Netflix dataset,
which has 100~million known ratings
from 480~thousand users for 18~thousand movies,
and we factorize the rating matrix with a rank of~500.

\textbf{Training methodology and performance metrics.}
Unless otherwise specified,
we train the image classification and video classification models
using the standard SGD algorithm with momentum,
and shuffle the training data every \emph{epoch}
(i.e., a whole pass over the training data).
The gradients of each training worker are normalized
with the training batch size before sending to the parameter server,
where the learning rate and momentum are applied.
For those supervised classification tasks,
the quality of the trained model is
defined as the classification accuracy
on a set of validation data,
and our experiments will focus on both
the convergence time and the converged validation accuracy
as the performance metrics.
Generally, users will need to specify the convergence condition,
and in our experiments,
we followed the common practice of other ML practitioners,
which is to test the validation accuracy every epoch
and consider the model as converged when the validation accuracy
plateaus (i.e., does not increase any more)~\cite{resnet,alexnet,inception-v3}.
Because of the noisiness of the validation accuracy traces,
we consider the ILSVRC12 and video classification benchmarks as converged
when the accuracy does not increase over the last 5~epochs,
and considered the \mbox{Cifar10} benchmark as converged
when the accuracy does not increase over the last 20~epochs.
Because \system trains the model for one more epoch after each re-tuning,
we configure \system to start re-tuning
one epoch before the model reaches the convergence condition
in order to be fair to the other setups.
Note that, even though the convergence condition
is defined in terms of the validation accuracy,
\system still evaluates tunable settings with the reported training loss,
because the training loss can be obtained every clock,
whereas the validation accuracy is only measured every epoch
(usually thousands of clocks for DNN training).

For the MF task, we define one clock as one whole pass over all training data,
without mini-batching.
Because MF is an unsupervised learning task,
we define its convergence condition as a fixed training loss value
(i.e., the model is considered as converged when it reaches that loss value),
and use the convergence time as a single performance metric,
with no re-tuning.
Based on guidance from ML experts
and related work using the same benchmark (e.g.,~\cite{gaia}),
we decided the convergence loss threshold as follows:
We first picked a relatively good tunable setting via grid search,
and kept training the model until the loss change was less
than 1\% over the last 10~iterations.
The achieved loss value is set as the convergence loss threshold,
which is 8.32$\times10^6$ for our MF setup.

\subsubsection{\System setup}
\label{sec:evaluation:tunable-setups}

\begin{table}
\centering
{\footnotesize
  \begin{tabular}{c|c}
    \hline
    {\bf Tunable} & {\bf Valid range} \\
    \hline
    Learning rate & \tabincell{c}{$10^{x}$, where $x \in [-5, 0]$} \\
    \hline
    Momentum & \tabincell{c}{DNN apps: $x \in [0.0, 1.0]$\\
                             Matrix factorization: N/A} \\
    \hline
    \tabincell{c}{Per-machine\\batch size} & \tabincell{c}{\mbox{Inception}-BN/\mbox{GoogLeNet}: $x \in \{2,4,8,16,32\}$ \\
                                   \mbox{AlexNet}: $x \in \{4,16,64,256\}$ \\
                                   RNN: $x \in \{1\}$ \\
                                   Matrix factorization: N/A
                                   } \\
    \hline
    Data staleness & $x \in \{0,1,3,7\}$ \\
    \hline
\end{tabular}
}
\caption{Tunable setups in the experiments.}
\label{tab:tunables}
\end{table}

Table~\ref{tab:apps} summarizes the tunables to be tuned in our experiments.
The tunable value ranges (except for the batch size)
are the same for all benchmarks,
because we assume little prior knowledge from users
about the tunable settings.
The (per-machine) batch size ranges are different for each model,
decided based on the maximum batch size that can fit in the GPU memory.
For the video classification task,
we can only fit one video in a batch,
so the batch size is fixed to one.

Except for specifying the tunables,
\system does not require any other user configurations,
and we used the same default configurations
(e.g., HyperOpt as the tunable searcher
and 10~samples for downsampling noisy progress)
for all experiments.
An application reports its training loss as the training progress
to \system every clock.

\subsubsection{Training system and cluster setup}

For the deep neural network experiments,
we use \mbox{GeePS}~\cite{geeps} connected with Caffe~\cite{caffe}
as the training system,
running distributed on 8~GPU machines (8~ML workers + 8~server shards).
Each machine has one NVIDIA Titan~X GPU,
with 12~GB of GPU device memory.
In addition to the GPU, each machine has one
E5-2698Bv3 Xeon CPU (2.0~GHz, 16~cores with 2~hardware threads each)
and 64~GB of RAM,
running 64-bit Ubuntu~16.04,
CUDA toolkit~8.0, and cuDNN~v5.
The machines are inter-connected via 40~Gbps Ethernet.

The matrix factorization experiments
use \mbox{IterStore}~\cite{iterstore} as the training system,
running distributed on 32~CPU machines (32~ML workers~+~32~server shards).
Each machine has four quad-core AMD Opteron 8354 CPUs
(16 physical cores in total) and 32~GB of RAM,
running 64-bit Ubuntu~14.04.
The machines are inter-connected via 20~Gb Infiniband.

\subsection{\system vs. state-of-the-art auto-tuning approaches}
\label{sec:evaluation:traditional-tuning-approaches}

This section experimentally compares our \system approach with
the state-of-the-art hyperparameter tuning approaches,
\mbox{Spearmint}~\cite{spearmint} and \mbox{Hyperband}~\cite{hyperband}.
To control for other performance factors,
we implemented the tuning logics
of those state-of-the-art approaches in our \system system.
All setups tune the same four tunables listed in Table~\ref{tab:tunables}.
The \mbox{Spearmint} approach
samples tunable settings with the Bayesian optimization
algorithm and trains the model to completion to evaluate each tunable
setting.\footnote{We used \mbox{Spearmint}'s open-sourced Bayesian optimization
implementation from \url{https://github.com/HIPS/\mbox{Spearmint}}
as of September 14, 2016.}
For the \mbox{Hyperband} approach,
we followed the ``\mbox{Hyperband} (Infinite horizon)'' algorithm~\cite{hyperband},
because the total number of epochs for the model to converge is unknown.
The Infinite horizon \mbox{Hyperband} algorithm
starts the searching with a small budget and doubles the budget over time.
For each given budget,
\mbox{Hyperband} samples tunable settings randomly from the search space,
and every few iterations, it will stop the half of configurations being
tried that have lower validation accuracies.


\begin{figure}[ht]
\centering
\includegraphics[keepaspectratio=1,width=1\columnwidth]{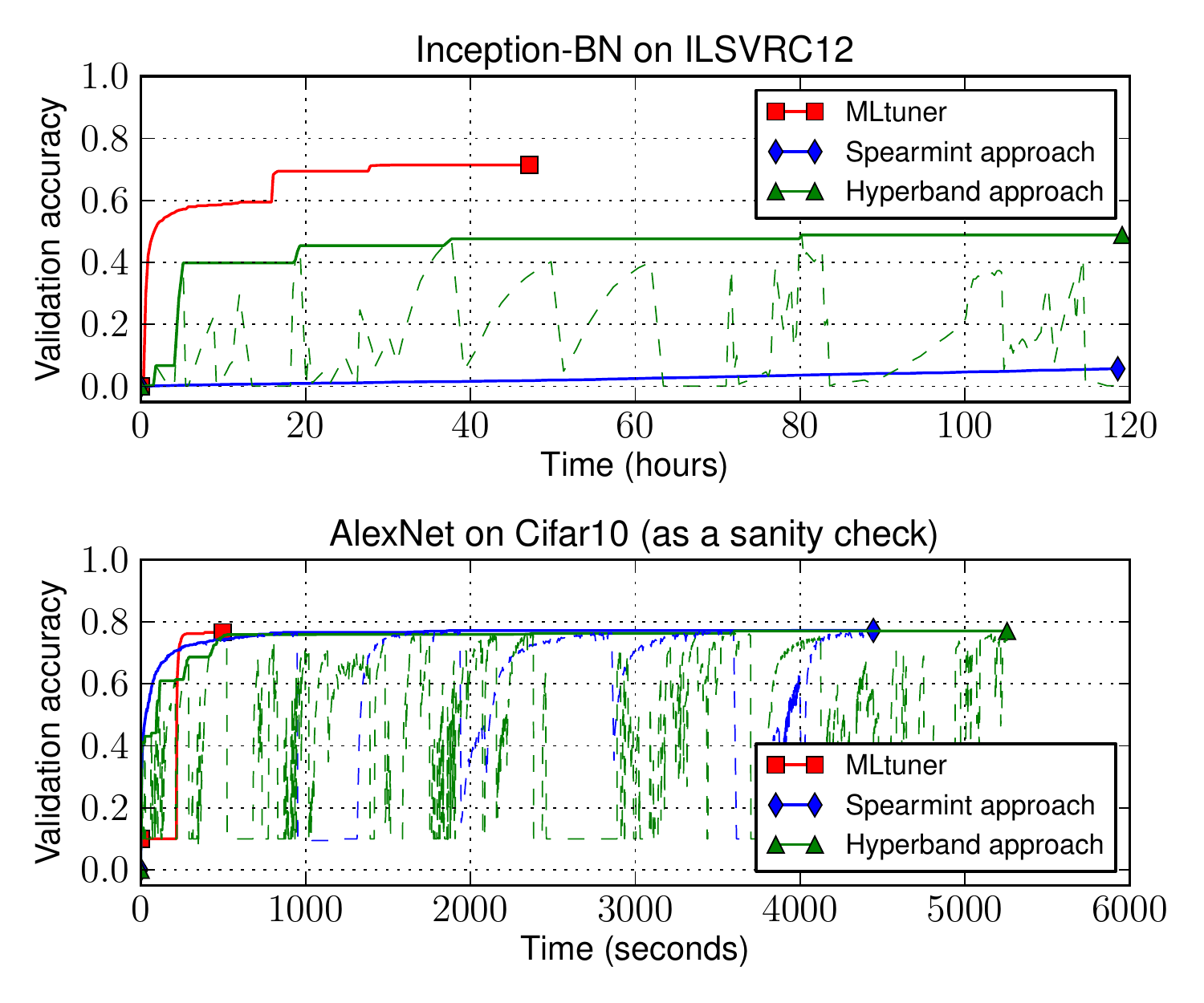}
\vspace{-.1in}
\caption{Runtime and accuracies of \system and the state-of-the-art approaches.
\small{
For \mbox{Spearmint} and \mbox{Hyperband},
the dashed curves show the accuracies of each configuration tried,
and the bold curves show the maximum accuracies achieved over time.}}
\label{fig:compare-baselines}
\end{figure}


Figure~\ref{fig:compare-baselines} shows
the runtime and achieved validation accuracies
of \mbox{Inception}-BN on ILSVRC12 and \mbox{AlexNet} on \mbox{Cifar10}.
For the larger ILSVRC12 benchmark,
\System performs much better than \mbox{Hyperband} and \mbox{Spearmint}.
After 5~days, \mbox{Spearmint} reached only 6\% accuracy,
and \mbox{Hyperband} reached only 49\% accuracy,
while \system converged to 71.4\% accuracy in just 2~days.
The \mbox{Spearmint} approach performs so badly
because the first tunable setting that it samples
sets all tunables to their minimum values
(learning rate=1e-5, momentum=0, batch size=2, data staleness=0),
and the small learning rate and batch size
cause the model to converge at an extremely slow rate.
We have tried running \mbox{Spearmint} multiple times,
and found their Bayesian optimization algorithm always
proposes this setting as the first one to try.
We also show the results on the smaller \mbox{Cifar10} benchmark as a sanity check,
because previous hyperparameter tuning work only reports results
on this small benchmark.
For the \mbox{Cifar10} benchmark,
all three approaches converged to approximately the same validation accuracy,
but \system is 9$\times$ faster than \mbox{Hyperband}
and 3$\times$ faster than \mbox{Spearmint}.\footnote{Since \mbox{Spearmint} and \mbox{Hyperband}
do not have stopping conditions of deciding when to quit the searching,
we measured the convergence time as
the time for each setup to reach 76\% validation accuracy.
If we set the stopping condition of \mbox{Spearmint}
as when the best 5~validation accuracies differ by less than 10\%,
\system finished the training in 90\% less time than \mbox{Spearmint}.}

\begin{figure}[ht]
\centering
\includegraphics[keepaspectratio=1,width=1\columnwidth]{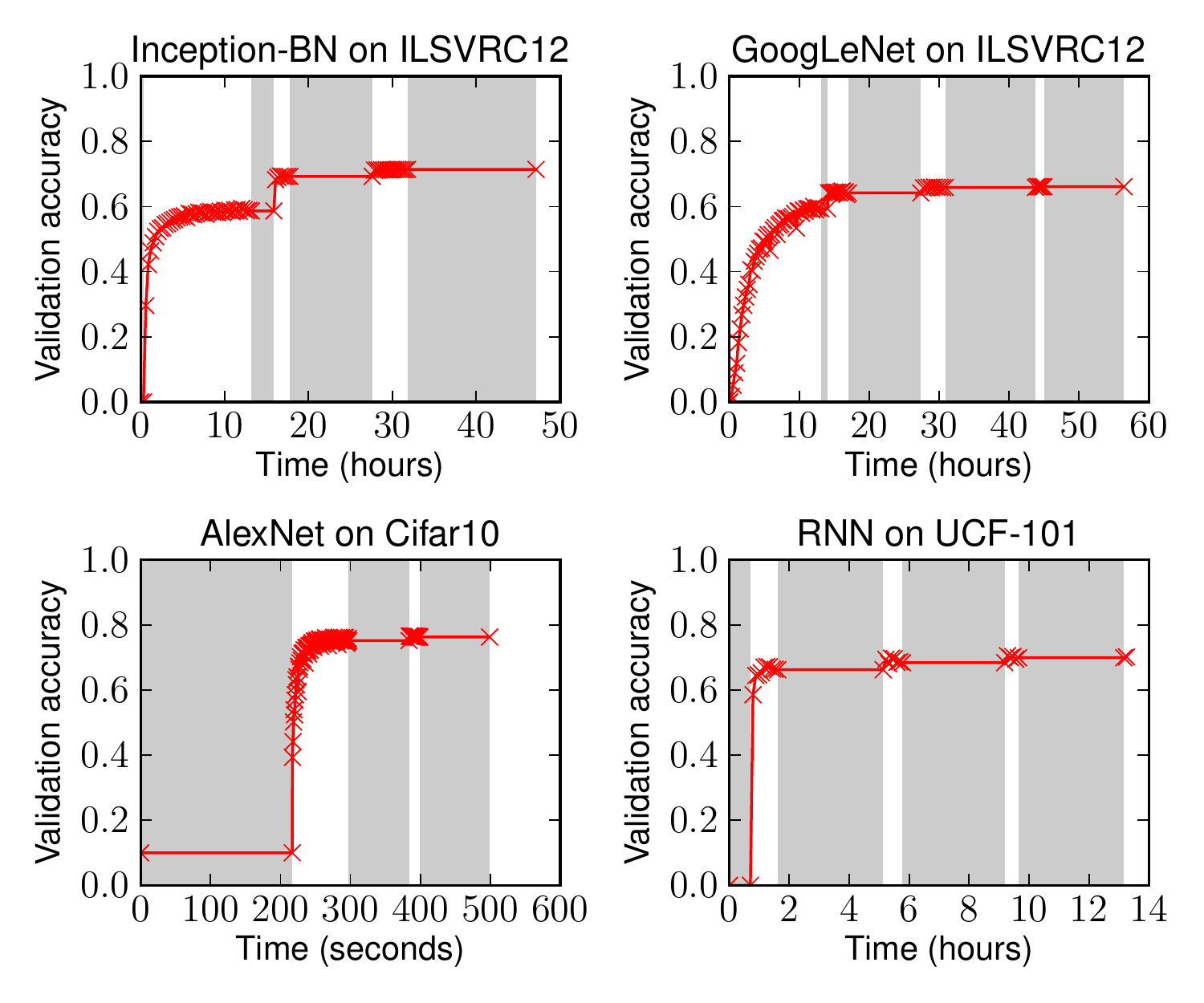}
\vspace{-.1in}
\caption{\System tuning/re-tuning behavior on four deep learning benchmarks.
\small{
The markers represent the validation accuracies measured at each epoch.
The shaded time ranges are when \system tunes/re-tunes tunables.}}
\label{fig:mltuner-accuracy}
\end{figure}

\begin{figure}[ht]
\centering
\includegraphics[keepaspectratio=1,width=1\columnwidth]{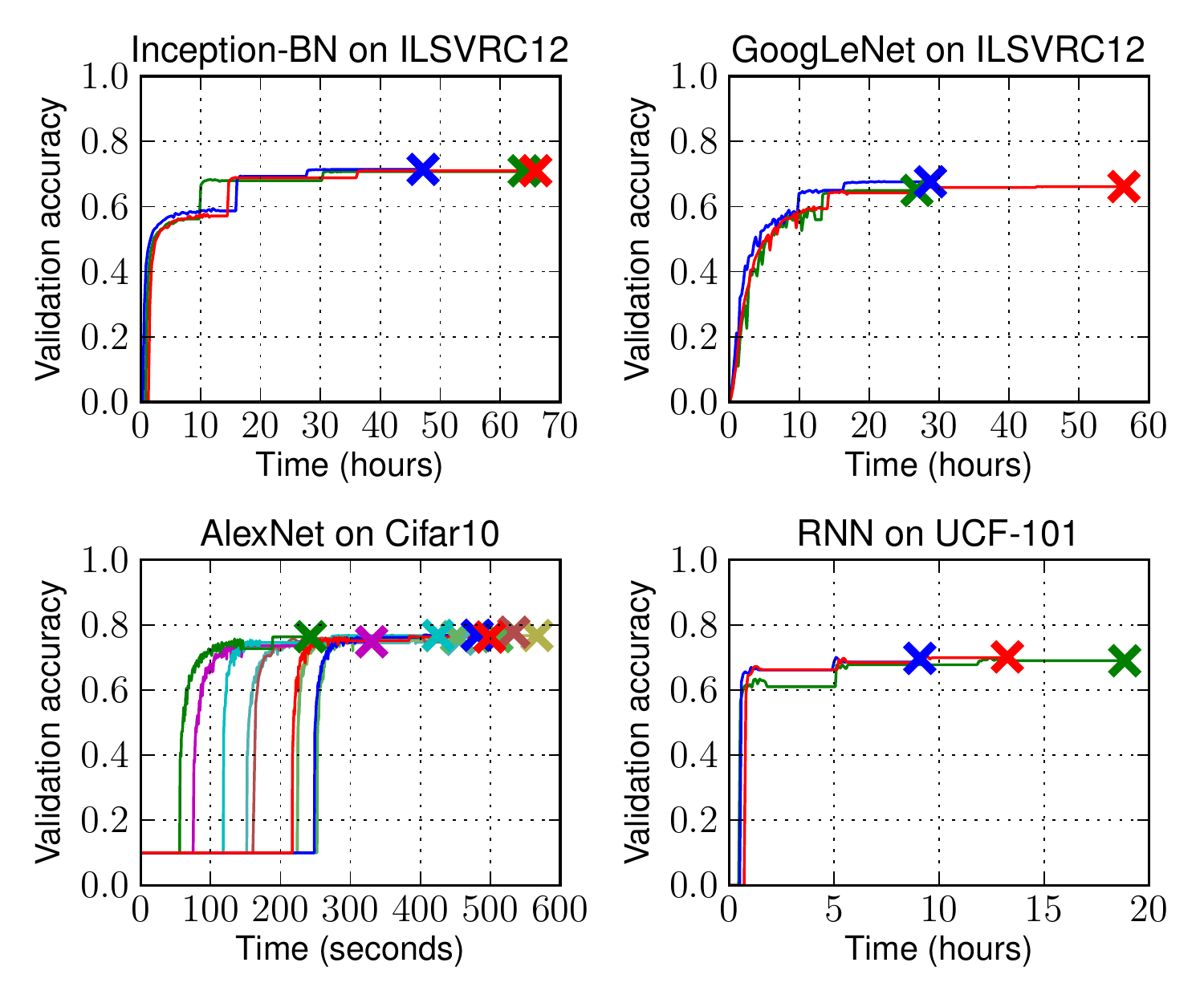}
\vspace{-.1in}
\caption{\System results of multiple runs.
\small{The larger ``x'' markers mark the end of each run.
The runs with the median performance are shown as the red curves in this figure
and as the ``\system'' curves in the other figures.}}
\label{fig:mltuner-accuracy-more-runs}
\end{figure}

Compared to previous approaches,
\system converges to much higher accuracies in much less time.
The accuracy jumps in the \system curves are caused by re-tunings.
Figure~\ref{fig:mltuner-accuracy} gives a more detailed view
of \system's tuning/re-tuning behavior.
\System re-tunes tunables when the validation accuracy plateaus,
and the results shows that the accuracy
usually increases after the re-tunings.
This behavior echoes experts' findings that,
when training deep neural networks,
it is necessary to change (usually decrease) the learning rate
during training, in order to get good validation accuracies
~\cite{inception-v3, inception-bn, googlenet, alexnet, resnet, poseidon}.
For the larger ILSVRC12 and RNN benchmarks,
there is little overhead (2\% to 6\%) from the initial tuning stage,
but there is considerable overhead from re-tuning,
especially from the last re-tuning, when the model has already converged.
That is because \system assumes no knowledge of the optimal model accuracy that
it is expected to achieve.
Instead, it automatically finds
the best achievable model accuracy via re-tuning.

Figure~\ref{fig:mltuner-accuracy-more-runs}
shows the \system results of multiple runs
(10~runs for Cifar10 and 3~runs each for the other benchmarks).
For each benchmark, \system consistently converges to nearly the same
validation accuracy.
The number of re-tunings and convergence time
are different for different runs.
This variance is caused by the
randomness of the \mbox{HyperOpt} algorithm used by \system,
as well as the inherent behavior of floating-point arithmetic
when the values to be reduced arrive in a non-deterministic order.
We observe similar behavior when not using \system, due to the
latter effect, which is discussed more
in Section~\ref{sec:eval-compare-with-manual}
(e.g., see Figure~\ref{fig:variance-test}).

\subsection{Tuning initial LR for adaptive LR algorithms}
\label{sec:eval-sgd-lr-tuning}

As we have pointed out in Section~\ref{sec:background-sgd-lr-tuning},
the adaptive learning rate tuning algorithms,
including \mbox{AdaRevision}~\cite{adarevision}, \mbox{RMSProp}~\cite{rmsprop},
\mbox{Nesterov}~\cite{nesterov}, \mbox{Adam}~\cite{adam},
\mbox{AdaDelta}~\cite{adadelta}, and \mbox{AdaGrad}~\cite{adagrad},
still require users to pick the initial learning rate.
This section will show that the initial learning rate settings
of those adaptive LR algorithms
still greatly impact the converged model quality and convergence time,
and that \system can be used to tune the initial learning rate for them.
For this set of experiments, \system only tunes the initial learning rate,
and does not re-tune,
so that \system will not affect the behaviors of the adaptive LR algorithms.

\subsubsection{Tuning initial LR improves solution quality}
\label{sec:eval-tune-initial-lr-for-accuracy}

\begin{figure}[ht]
\centering
\includegraphics[keepaspectratio=1,width=1\columnwidth]{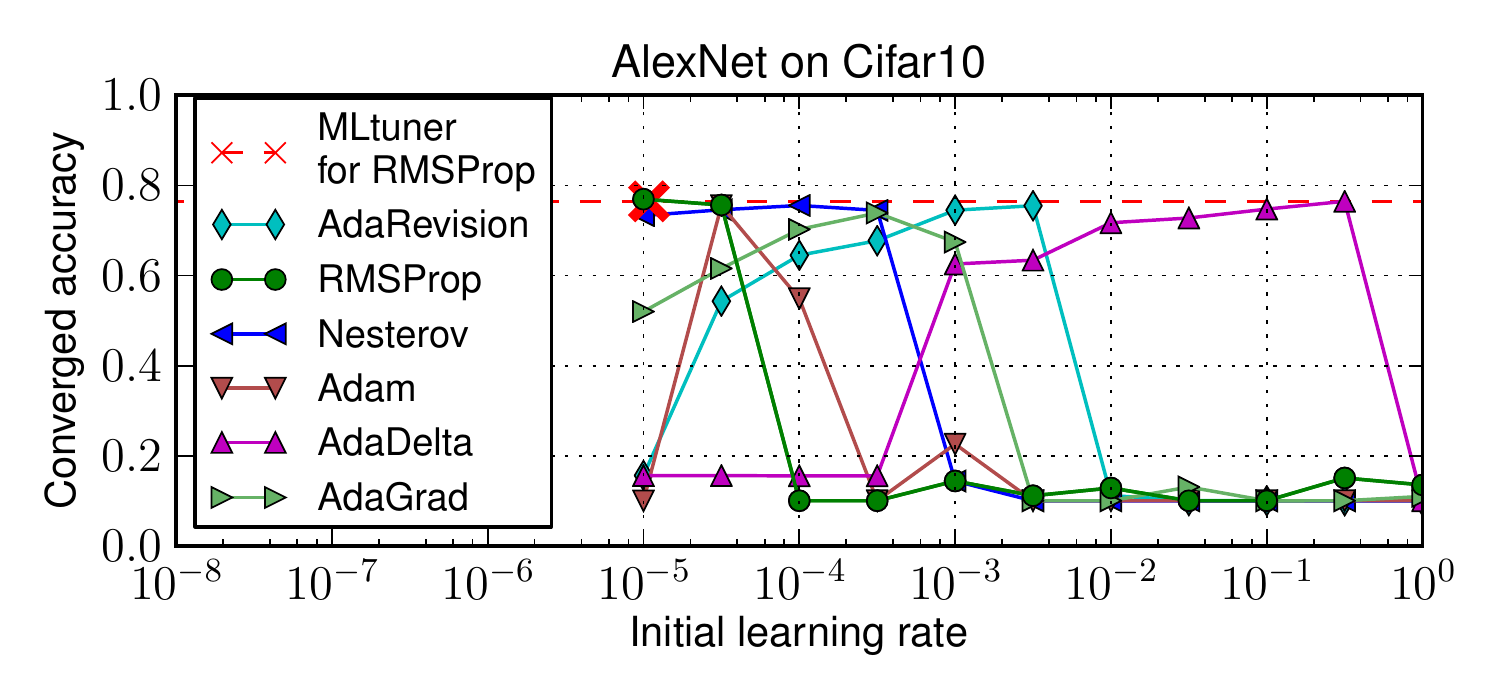}
\vspace{-.2in}
\caption{Converged validation accuracies
when using different initial learning rates.
\small{The ``x'' marker marks the LR picked by \system{} for RMSProp.}}
\label{fig:all-lrs-cifar}
\end{figure}

Figure~\ref{fig:all-lrs-cifar} shows the converged validation accuracies
of \mbox{AlexNet} on \mbox{Cifar10}
with different adaptive LR algorithms
and different initial learning rate settings.
We used the smaller \mbox{Cifar10} benchmark,
so that we can afford to train the model to convergence
with many different initial LR settings.
For the other tunables,
we used the common default values
(momentum=0.9, batch size=256, data staleness=0)
that are frequently suggested in the
literature~\cite{alexnet,googlenet,inception-bn,geeps}.
The results show that
the initial LR setting
greatly affects the converged accuracy,
that the best initial LR settings differ across adaptive LR algorithms,
and that the optimal accuracy for a given algorithm can only be achieved
with one or two settings in the range.
The result also shows that \system can effectively pick good initial LRs
for those adaptive LR algorithms,
achieving close-to-ideal validation accuracy.
The graph shows only the tuning result for \mbox{RMSProp}
because of limited space,
but for all the 6~adaptive LR algorithms, the accuracies achieved by \system
differ from those with the optimal setting by less than 2\%.

\subsubsection{Tuning initial LR improves convergence time}

\begin{figure}[ht]
\centering
\vspace{-.1in}
\includegraphics[keepaspectratio=1,width=0.8\columnwidth]{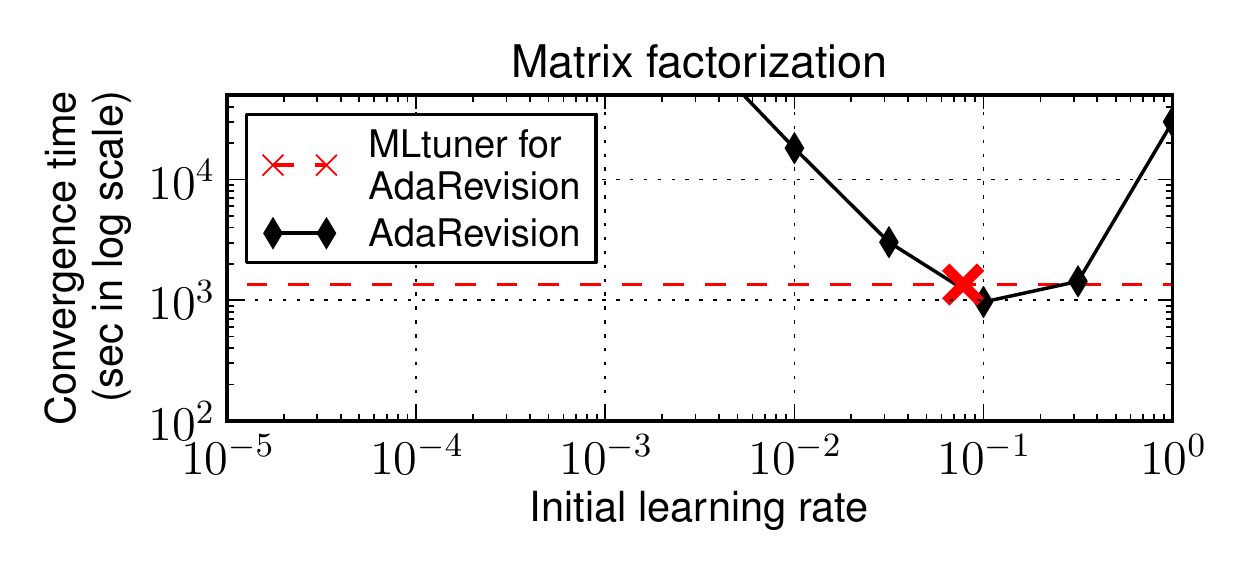}
\vspace{-.1in}
\caption{Convergence time when
using different initial learning rates.
\small{The ``x'' marker marks the LR picked by \system.}}
\label{fig:all-lrs-mf}
\end{figure}

Figure~\ref{fig:all-lrs-mf} shows the convergence time
when using different initial \mbox{AdaRevision} learning rate settings,
for the matrix factorization benchmark.
Because the model parameters of the MF task have uneven update frequency,
practitioners often use \mbox{AdaRevision}~\cite{adarevision}
to adjust its per-parameter learning rates~\cite{bosen}.
Among all settings, more than 40\% of them caused
the model to converge over an order of magnitude slower
than the optimal setting.
We also show that,
when tuning the initial LR with \system,
the convergence time (including the \system tuning time)
is close to ideal
and is over an order of magnitude faster than
leaving the initial LR un-tuned.

\subsection{\System vs. idealized manually-tuned settings}
\label{sec:eval-compare-with-manual}

\begin{figure}[ht]
\centering
\includegraphics[keepaspectratio=1,width=1\columnwidth]{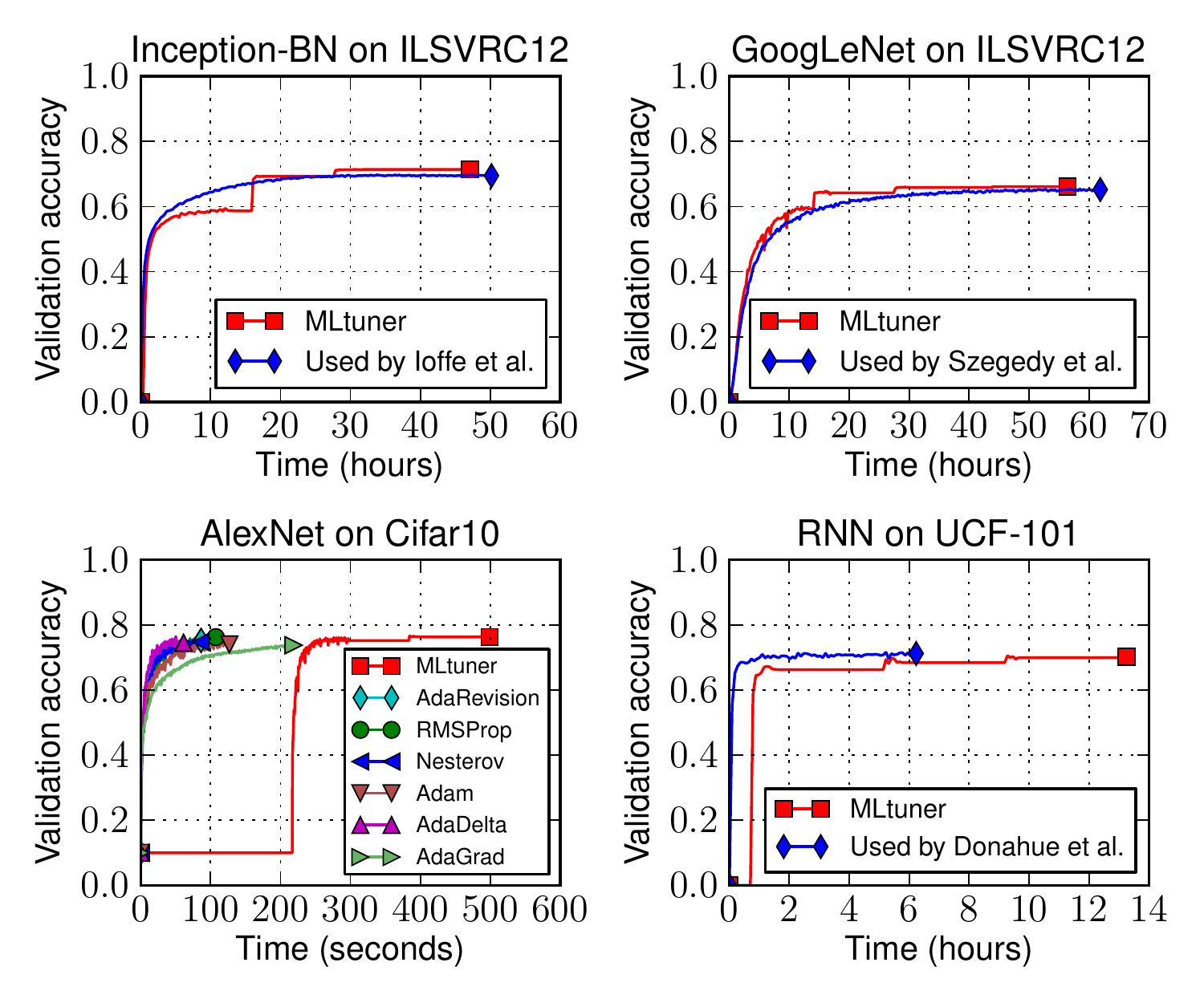}
\vspace{-.1in}
\caption{\System compared with manually tuned settings.
\small{For comparison purpose, we have run the manually tuned settings
(except for \mbox{Cifar10})
for long enough to ensure that their accuracies will not increase any more,
rather than stopping them according to the convergence condition.}
}
\label{fig:vs-best-settings}
\end{figure}

Figure~\ref{fig:vs-best-settings} compares the performance of \system,
automatically tuning all four tunables listed in Table~\ref{tab:tunables},
with an idealized ``manually tuned'' configuration of the tunable settings.
The intention is to evaluate \System{}'s overhead relative to what
an expert who already figured out the best settings (e.g., by extensive
previous experimentation) might use.
For the \mbox{Cifar10} benchmark,
we used the optimal initial LRs
for the adaptive algorithms,
found via running all possible settings to completion,
and used effective default values for the other tunables
(m=0.9, bs=256, ds=0).
\cui{TODO: probably need to add a curve where no adaptive LR algorithm is used.}
The results show that, among all the adaptive algorithms,
\mbox{RMSProp} has the best performance.
Compared to the best \mbox{RMSProp} configuration,
\system reaches the same accuracy,
but requires about 5$\times$ more time.

For the other benchmarks,
our budget does not allow us to run all the possible settings to completion
to find the optimal ones.
Instead, we compared with manually tuned settings
suggested in the literature.
For \mbox{Inception}-BN,
we compared with the manually tuned setting suggested by Ioffe et al.
in the original \mbox{Inception}-BN paper~\cite{inception-bn},
which uses an initial LR of 0.045
and decreases it by 3\% every epoch.
For \mbox{GoogLeNet},
we compared with the manually tuned setting suggested by Szegedy et al.
in the original \mbox{GoogLeNet} paper~\cite{googlenet},
which uses an initial LR of 0.0015,
and decreases it by 4\% every 8~epochs.
For RNN,
we compared with the manually tuned setting
suggested by Donahue et al.~\cite{lrcn},
which uses an initial LR of 0.001,
and decreases it by 7.4\% every epoch.
\footnote{
\cite{lrcn} does not specify the tunable settings,
but we found their settings in their released source code
at \url{https://github.com/LisaAnne/lisa-caffe-public}
as of April 16, 2017.
}
All those manually tuned settings set momentum=0.9 and data staleness=0,
and \mbox{Inception}-BN and \mbox{GoogLeNet} set batch size=32.
Compared to the manually tuned settings,
\system achieved the same accuracies for \mbox{Cifar10} and RNN,
and higher accuracies for \mbox{Inception}-BN (71.4\% vs. 69.8\%)
and \mbox{GoogLeNet} (66.2\% vs. 64.4\%).
The higher \system accuracies might be because of two reasons.
First, those reported settings were tuned for
potentially different hardware setups (e.g., number of machines),
so they might be suboptimal for our setup.
Second, those reported settings used fixed learning rate decaying rates,
while \system is more flexible and can use any learning rate via re-tuning.

As expected, \System requires more time to train than when an expert
knows the best settings to use.
The difference is 5$\times$ for the small \mbox{Cifar10} benchmark, but is much
smaller for the larger ILSVRC12 benchmarks, because the tuning times
are amortized over much more training work.
We view these results to be very positive, since knowing the ML task specific
settings traditionally requires extensive experimentation that
significantly exceeds \System{}'s overhead or even the much higher overheads
for previous approaches like \mbox{Spearmint} and \mbox{Hyperband}.


\begin{figure}[ht]
\centering
\vspace{-.1in}
\includegraphics[keepaspectratio=1,width=1\columnwidth]{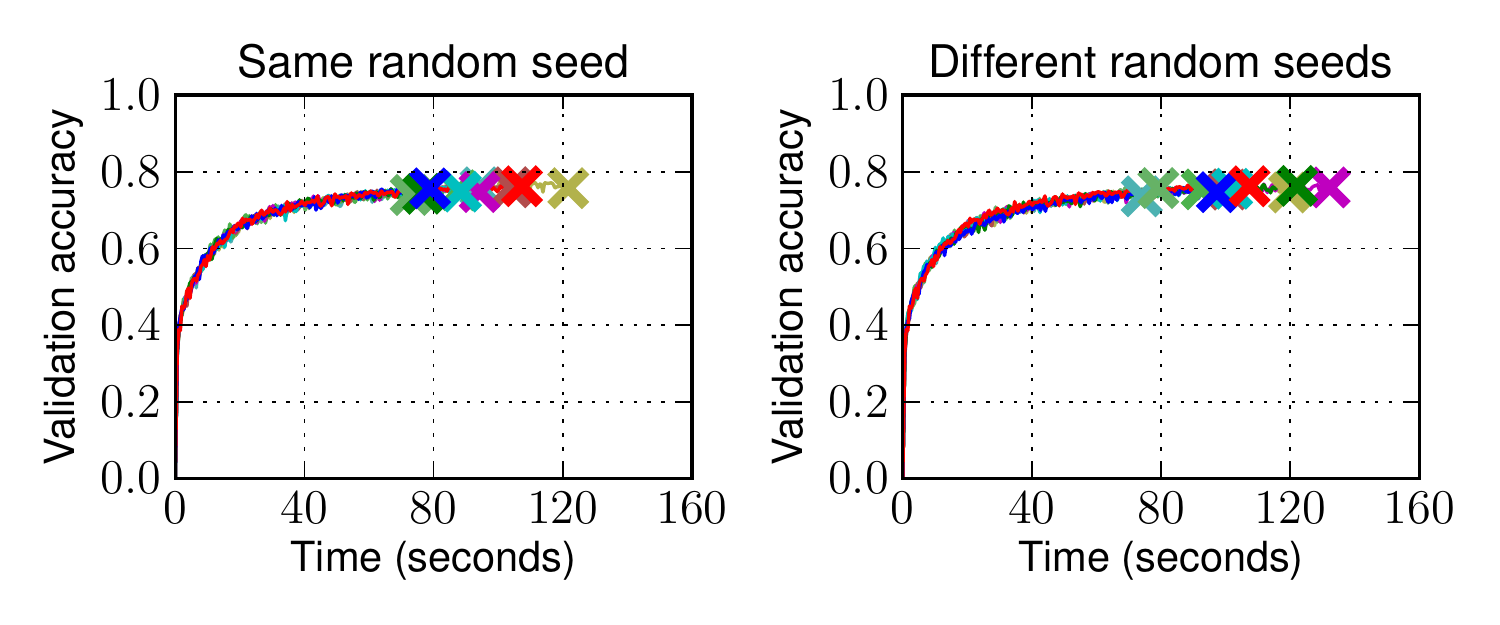}
\vspace{-.2in}
\caption{Performance for multiple training runs of AlexNet on Cifar10
with \mbox{RMSProp} and the optimal initial LR setting.}
\label{fig:variance-test}
\end{figure}

Figure~\ref{fig:variance-test} shows training performance for multiple runs
of \mbox{AlexNet} with \mbox{RMSProp} using the same (optimal) initial LR.
In the left graph, all runs initialize model parameters
and shuffle training data with the same random seed.
In the right graph, a distinct random seed is used for each run.
We did 10~runs for each case
and stopped each run when it reached the convergence condition.
The result shows considerable variation in their convergence times across runs,
which is caused by random initialization of parameters,
training data shuffling,
and non-deterministic order of floating-point arithmetic.
The coefficients of variation
(CoVs~=~standard deviation divided by average) of their convergence times
are 0.16 and 0.18, respectively,
and the CoVs of their converged accuracies
are both 0.01.
For the 10~\system runs on the same benchmark shown in
Figure~\ref{fig:mltuner-accuracy-more-runs},
the CoV of the convergence time is 0.22,
and the CoV of the converged accuracy is 0.01.




\subsection{Robustness to suboptimal initial settings}

\begin{figure}[ht]
\centering
\vspace{-.1in}
\includegraphics[keepaspectratio=1,width=1\columnwidth]{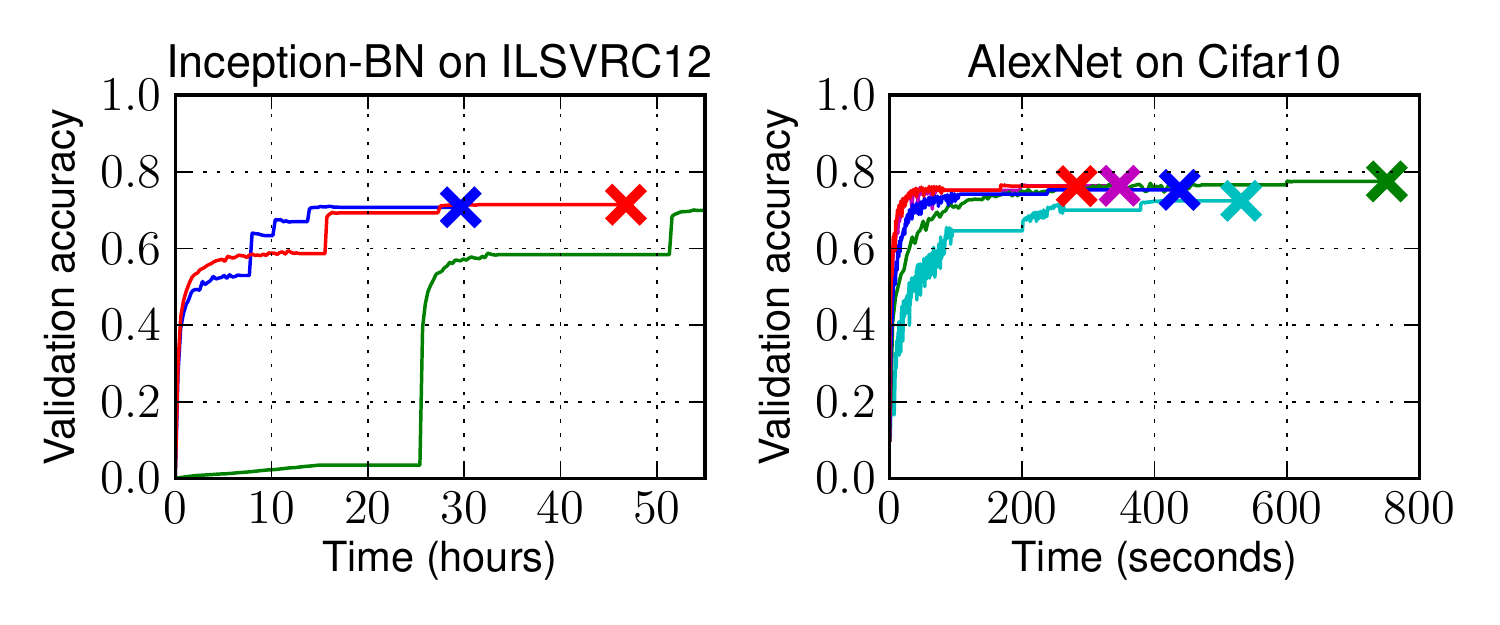}
\vspace{-.2in}
\caption{\System performance with hard-coded initial tunable settings.
\small{The red curves used the tuned initial settings,
and the other curves used randomly selected suboptimal initial settings.
}}
\label{fig:start-with-bad}
\end{figure}

This set of experiments studies the robustness of \system.
In particular, we turned off the initial tuning stage of \system
and had \system use a hard-coded suboptimal tunable setting (picked randomly)
as the initial setting.
The result in Figure~\ref{fig:start-with-bad}
shows that, even with suboptimal initial settings,
\system is still able to robustly converge to
good validation accuracies via re-tuning.

\subsection{Scalability with more tunables}

\begin{figure}[ht]
\centering
\vspace{-.1in}
\includegraphics[keepaspectratio=1,width=0.7\columnwidth]{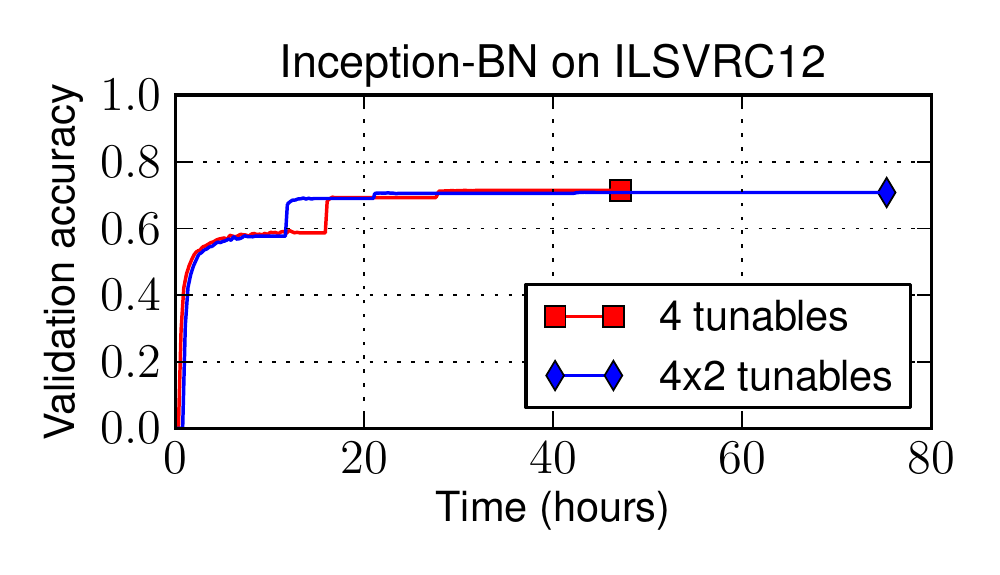}
\vspace{-.1in}
\caption{\System performance with more tunables.}
\label{fig:inception-dup-tunables}
\end{figure}

Figure~\ref{fig:inception-dup-tunables} shows
\system's scalability to the number of tunables.
For the ``4$\times$2 tunables'' setup,
we duplicated the 4~tunables listed in Table~\ref{tab:tunables},
making it a search space of 8~tunables.
Except for making the search space larger,
the added 4~tunables are transparent to the training system and
do not control any other aspects of the training.
The result shows that,
with 8~tunables to be tuned,
\system still successfully converges to the same validation accuracy.
The tuning time increases by about 2$\times$,
which is caused by
the increased number of settings tried by HyperOpt
before it reaches the stopping condition.

\section{Conclusions}

\system automatically tunes the training tunables that can have major impact
on the performance and effectiveness of ML applications.
Experiments with three real ML applications on two real ML systems
show that \system has robust performance and
outperforms state-of-the-art auto-tuning approaches
by over an order of magnitude on large problems.
\System also automatically achieves performance comparable to
manually-tuned settings by experts.

\newpage

%
%
%





{
\bibliography{main.bib}
\bibliographystyle{acm}
}




\end{document}